\documentclass[runningheads]{llncs}
\usepackage{graphicx}
\usepackage{multirow}
\usepackage{todonotes}
\usepackage{xcolor,colortbl}
\usepackage{amssymb}
\usepackage{amsmath}
\usepackage{graphics}

\begin{document}
\title{Sequence-to-Sequence Models for Extracting Information from Registration and Legal Documents}
\titlerunning{Seq2Seq Models for Extracting Information from Documents}
\author{Ramon Pires\inst{1,2}\orcidID{0000-0002-0023-1971} \and
Fábio C. de Souza\inst{1,3}\orcidID{0000-0003-0174-4506} \and 
Guilherme Rosa\inst{1,3} \and 
Roberto A. Lotufo\inst{1,3}\orcidID{0000-0002-5652-0852} \and \\
Rodrigo Nogueira\inst{1,3}\orcidID{0000-0002-2600-6035}
}

\authorrunning{R. Pires et al.}

\institute{NeuralMind Inteligência Artificial, São Paulo, Brazil
\email{\{ramon.pires,fabiosouza,guilherme.rosa,roberto,rodrigo.nogueira\}@neuralmind.ai}\\
Institute of Computing, University of Campinas, Campinas, Brazil \and
School of Electrical and Computer Engineering, University of Campinas, Campinas, Brazil
}


\maketitle              

\begin{abstract}
A typical information extraction pipeline consists of token- or span-level classification models coupled with a series of pre- and post-processing scripts. In a production pipeline, requirements often change, with classes being added and removed, which leads to nontrivial modifications to the source code and the possible introduction of bugs. In this work, we evaluate sequence-to-sequence models as an alternative to token-level classification methods for information extraction of legal and registration documents. 
We finetune models that jointly extract the information and generate the output already in a structured format.
Post-processing steps are learned during training, thus eliminating the need for rule-based methods and simplifying the pipeline.
Furthermore, we propose a novel method to align the output with the input text, thus facilitating system inspection and auditing.
Our experiments on four real-world datasets show that the proposed method is an alternative to classical pipelines.
\end{abstract}

\section{Introduction}
\label{sec:introduction}

Current commercial information extraction (IE) systems consist of individual modules organized in a pipeline with multiple branches and merges controlled by manually defined rules. Most of such modules are responsible for extracting or converting a single piece of information from an input document~\cite{sarawagi2008information,chieze2010automatic,bommarito2021lexnlp}.
In production pipelines, the requirements and specifications often change, with new types of documents being added and more information being extracted. This leads to higher maintenance costs due to an ever larger number of individual components.

Although recent pretrained language models significantly improved the effectiveness of IE systems, modern pipelines are made of various of such models, each responsible for extracting or processing a small portion of the document.
Due to the large computational demands of these models, it can be challenging to deploy a low-latency low-cost IE pipeline.

\begin{figure}
     \centering
     \includegraphics[width=\textwidth]{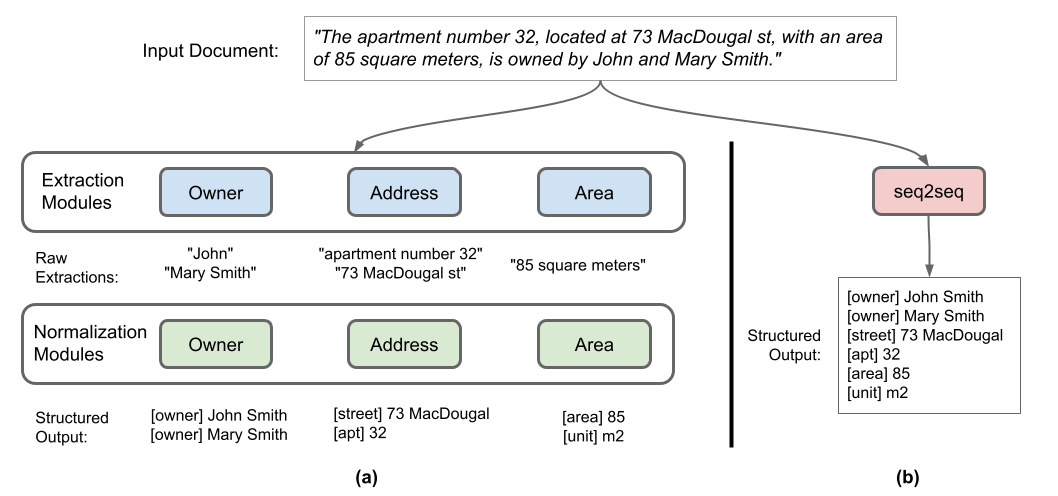}
     \vspace{-0.8cm}
     \caption{\textbf{a)} A classical IE pipeline consisting of multiple extraction and normalization modules. \textbf{b)} Our proposed IE framework that converts the input document into structured data using a single seq2seq model.}
    \label{fig:overview}
\end{figure}

In this work, we study the viability of a framework for information extraction based on a single sequence-to-sequence (seq2seq) model finetuned on a question answering (QA) task, for extracting and processing information from legal and registration documents.
The main advantage of this framework is that a single model needs to be trained and maintained. Thus, it can be shared by multiple projects with different requirements and types of documents. Figure~\ref{fig:overview} depicts an illustration comparing a classical IE pipeline with the proposed one.

We validate this single-model framework on four datasets that represent demands from real customers.
We summarize our contributions as follows:
\begin{itemize}
    \item We show that a single seq2seq model is a competitive alternative to a classical pipeline made of multiple extraction and normalization modules (see Section~\ref{sec:ablation} for an ablation study). Our model is trained end-to-end to generate the structured output in a single decoding pass.
    \item We evaluate different pretraining and intermediate finetuning strategies and show that target language tokenization and pretraining are by far the most important contributors to the final performance. Further pretraining on the legal domain and finetuning on a similar task have minor contributions.
    \item One of the limitations of generative models is that the location of extracted tokens in the input text can be ambiguous. We propose a novel method to address this problem (Section~\ref{sec:sent_ids_canonical}).
\end{itemize}

\section{Related Work}
\label{sec:related-work}

A go-to choice of methods for information extraction are token-level classification models, which predict a label for each element in the input sequence~\cite{chieu2003named,carreras2003learning,florian2003named,lample2016neural,chiu2016named,baevski2019cloze,li2020dice,chen2019bert}.
Seq2seq models are typically used to solve complex language tasks like machine translation, text summarization, and question answering~\cite{lewis2020bart,song2019mass,dong2019unified,zhang2020pegasus}. This special class of language models has shown excellent performance on tasks that are commonly tackled by token-level or span-level classification methods. For instance, T5~\cite{raffel2019exploring} is a seq2seq model that achieves the state of the art on various extractive question-answering and natural language inference tasks.
Others have confirmed that this single model can replace multiple task-specific NLP models~\cite{du2021all}.

We are not the first to apply seq2seq models to IE tasks.
Chen et al.~\cite{chen2018learning} proposed a seq2seq model that can recognize previously unseen named entities.
Athiwaratkun et al.~\cite{athiwaratkun2020augmented} finetuned a T5 model to jointly learn multiple tasks such as named entity recognition (NER), slot labeling and intent classification.
Hybrids of token-level classification and seq2seq models were also used to extract information from legal documents~\cite{de2020legal}.
Others use a seq2seq model for nested and discontinuous NER tasks~\cite{strakova-etal-2019-neural,yan2021unified}.
Most of the datasets used in these works contain shorter input texts than the ones used here. 

Although IE systems have a long history in the scientific literature, there are few studies that analyze their use in commercial NLP pipelines.
There are legal NER datasets similar to the ones used in this work~\cite{de2018lener,angelidis2018named,leitner2020dataset,huang2020named}, but they rarely reflect the complexities of production pipelines, such as processing scanned documents with OCR errors and extracting nested and discontiguous entities.

\section{Methodology}
\label{sec:methodology}

\begin{figure}[!ht]
    \centering
    \includegraphics[width=\textwidth]{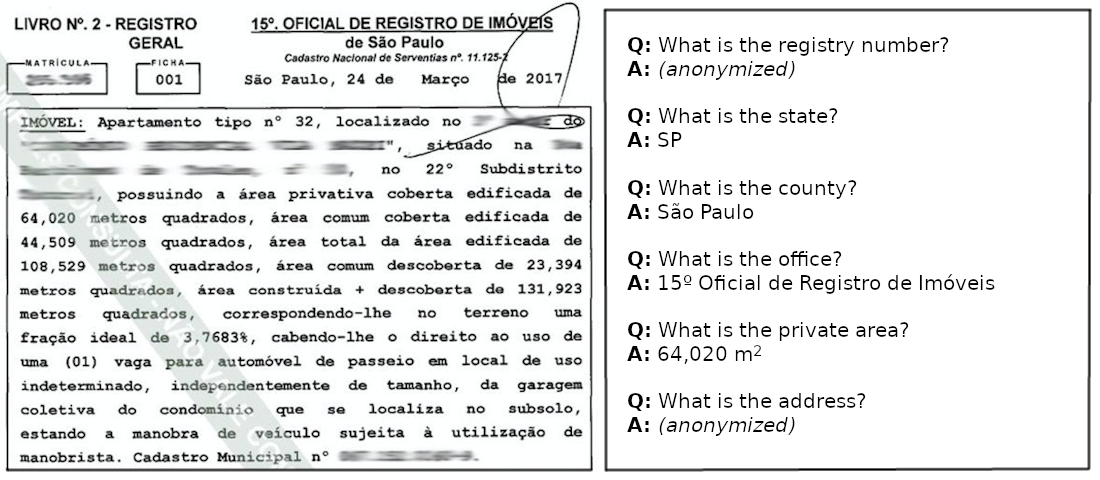}
    \vspace{-0.5cm}
    \caption{Example of Registry of Property (left) along with its questions and answers annotated by a human (right). Some answers are omitted to preserve anonymity.
    }
    \label{fig:documents}
\end{figure}

This section first describes input and output formats used by our models. Then, we present compound QAs, and introduce sentence IDs and answers in canonical (normalized) format. Figure~\ref{fig:documents} depicts one example of a Registry of Property, along with some human-annotated questions and answers.

\vspace{-0.5cm}

\subsection{Questions and Answers}
\label{sec:questions_and_answers}

Our method uses questions coupled with contexts as input and answers as output. In this section, we describe how we prepare QAs and contexts for extracting information from fields of different legal and registration documents.




We seek to format the answers by adding clues for each category of information. Clues are indispensable for designing compound answers (see Section~\ref{sec:comp_qas}). To maintain consistency and standardization, those clues are reused for related fields, regardless of the document type. We format the answers by preceding the core answer with the clue in square brackets followed by a colon. 



Due to the quadratic cost with respect to the sequence length of transformer models, we have to divide long documents into windows (e.g., of 10 sentences) with a certain overlap and repeat the questions in each window.
In some cases, the model should not provide an answer because the necessary information is not present in the current window. Thus, we finetune it to output ``N/A'' (an abbreviation for not available) for those question and context pairs.
At inference time, the extracted information is the one that achieves the highest probability among all windows, except for N/A responses. The extracted information is empty only if the model has generated ``N/A'' in all windows.

\subsection{Compound QAs}
\label{sec:comp_qas}


Some subsets of fields are often closely related or even appear connected. The classical pipeline is cumbersome as it requires the model to analyze each document oftentimes for extracting information of the same scope. An example is the \emph{address} field, whose extraction requires recovering, in general, seven individual subfields: street, number, complement, district, city, state, and zip code.

We propose a novel alternative for extracting all that information at once by using compound QAs.
We represent compound answers by preceding each field response with the respective field indicator (clue) in square brackets, and concatenating them all in a fixed, pre-set order (e.g., value and unit, in sequence). We also replace the set of questions by a single, more generic question. Table~\ref{tab:all_qas} presents examples of compound QAs (column \emph{Comp} marked). 

\begin{table}[!ht]
\caption{Examples of different formats of contexts, questions and answers. When the column \emph{Sent} is marked (\checkmark), question and answer pairs use the ``sent IDs'' context. When not marked, the ``original'' context is used.}\label{tab:all_qas}
\centering\resizebox{\textwidth}{!}{
\begin{tabular}{l p{11cm} c c c}
\hline
 & \multicolumn{4}{p{13.3cm}}{Contexts} \\
\hline
\parbox[t]{2mm}{\multirow{3}{*}{\rotatebox[origin=c]{90}{{\tiny original}}}} & \multicolumn{4}{|p{13.3cm}}{Apartment type nº 32, located on the 10th floor of the Central Building, situated at 1208 Santos Dumont St., having a private covered built area of 64,020 square meters, a common covered built area of 44,509 square meters...} \\ \hline 
\parbox[t]{2mm}{\multirow{4}{*}{\rotatebox[origin=c]{90}{{\tiny \ \ \ \ \  sent IDs}}}} & \multicolumn{4}{|p{13.3cm}}{[SENT1] Apartment type nº 32, [SENT2] located on the 10th floor of the Central Building, [SENT3] situated at 1208 Santos Dumont St., [SENT4] having a private covered built area of 64,020 square meters, [SENT5] a common covered built area of 44,509 square meters...} \\ 
\hline
 &  Questions and Answers & Comp & Sent & Raw \\
 \hline
\multirow{4}{*}{(1)} & {\tiny (Q)} What is the value of the private area? & \multirow{4}{*}{-} & \multirow{4}{*}{-} & \multirow{4}{*}{-} \\
                     & {\tiny (A)} [value]: 64.02 & & & \\ 
                     & {\tiny (Q)} What is the unit of the private area? & & & \\ 
                     & {\tiny (A)} [unit]: $m^2$ & & & \\
                     \hline
\multirow{2}{*}{(2)} & {\tiny (Q)} What is the private area? & \multirow{2}{*}{\checkmark} & \multirow{2}{*}{-} & \multirow{2}{*}{-}  \\
                     & {\tiny (A)} [value]: 64.02 [unit]: $m^2$ & & & \\
                     \hline
\multirow{4}{*}{(3)} & {\tiny (Q)} What is the value of the private area? & \multirow{4}{*}{-} & \multirow{4}{*}{\checkmark} & \multirow{4}{*}{-} \\
                     & {\tiny (A)} [SENT4] [value]: 64.02 & & & \\ 
                     & {\tiny (Q)} What is the unit of the private area? & & & \\ 
                     & {\tiny (A)} [SENT4] [unit]: $m^2$ & & & \\
                     \hline
\multirow{2}{*}{(4)} & {\tiny (Q)} What is the private area? & \multirow{2}{*}{\checkmark} & \multirow{2}{*}{\checkmark} & \multirow{2}{*}{-}  \\
                     & {\tiny (A)} [SENT4] [value]: 64.02 [SENT4] [unit]: $m^2$ & & & \\
                     \hline
\multirow{4}{*}{(5)} & {\tiny (Q)} What is the value of the private area and how does it appear in the text? & \multirow{4}{*}{-} & \multirow{4}{*}{-} & \multirow{4}{*}{\checkmark} \\
                     & {\tiny (A)} [value]: 64.02 [text] 64,020 & & & \\ 
                     & {\tiny (Q)} What is the unit of the private area and how does it appear in the text? & & & \\ 
                     & {\tiny (A)} [unit]: $m^2$ [text] square meters & & & \\
                     \hline
\multirow{2}{*}{(6)} & {\tiny (Q)} What is the private area and how does it appear in text? & \multirow{2}{*}{\checkmark} & \multirow{2}{*}{-} & \multirow{2}{*}{\checkmark}  \\
                     & {\tiny (A)} [value]: 64.02 [text] 64,020 [unit]: $m^2$ [text] square meters & & & \\
                     \hline
\multirow{4}{*}{(7)} & {\tiny (Q)} What is the value of the private area and how does it appear in the text? & \multirow{4}{*}{-} & \multirow{4}{*}{\checkmark} & \multirow{4}{*}{\checkmark} \\
                     & {\tiny (A)} [SENT4] [value]: 64.02 [text] 64,020 & & & \\ 
                     & {\tiny (Q)} What is the unit of the private area and how does it appear in the text? & & & \\ 
                     & {\tiny (A)} [SENT4] [unit]: $m^2$ [text] square meters & & & \\
                     \hline
\multirow{2}{*}{(8)} & {\tiny (Q)} What is the private area and how does it appear in text? & \multirow{2}{*}{\checkmark} & \multirow{2}{*}{\checkmark} & \multirow{2}{*}{\checkmark}  \\
                     & {\tiny (A)} [SENT4] [value]: 64.02 [text] 64,020 [SENT4] [unit]: $m^2$ [text] square meters & & & \\
                     \hline
\end{tabular}
}
\end{table}

Compound QAs can be valuable to extract discontinuous information (like nested NER)~\cite{strakova-etal-2019-neural,yan2021unified}. 
The user may request, for example, the complete list of heirs along with their respective registration numbers and the fraction. 
The use of compound QAs would be beneficial as it requires a single shot IE for each heir. 

\subsection{Sentence IDs and Canonical Format}
\label{sec:sent_ids_canonical}


Commercial IE systems are part of important decision making pipelines, and as such, they need to be constantly monitored and audited.
One way to monitor the quality of predictions is to know the location in the input text from which the information was extracted.
However, the location cannot be trivially inferred from the output of seq2seq models if the information appears more than once in the text. 
To address this limitation, we propose the use of \textit{sentinel tokens} that represent sentence IDs and allow the generative model to reveal the location of its prediction in the original sequence. Table~\ref{tab:all_qas} shows examples of QAs with sentence IDs (column \emph{Sent} marked). 

The questions are not changed for requesting sentence IDs. The document text, however, must be split into sentences following some criterion (e.g., paragraph, form field, number of tokens, lines). Then, each sentence is preceded by its own sentinel token using the pattern [SENT$x$], in which $x$ represents the sentence ID, starting from $1$ and increasing for each new sentence.
The answers are preceded by the sentinel token of the sentence it came from. If the answer is compound, each component has its sentence ID preceding the clue and response.


Often, certain types of information appear in a document in a variety of formats.
For example, a date may be expressed in a day-month-year format, such as ``20 November 2021", or abbreviated with some sort of separator like ``20/11/2021" or ``20-11-2021".
For a commercial system, it is required to extract those types of information in a standardized, canonical format to harmonize varied formats and ensure accuracy in all situations. Our IE system is able to directly extract those particular fields in the canonical format.

However, the canonical format might not match the span in the original text, and thus the use of sentence IDs may not be enough for locating the extracted information in the original document. To maintain this functionality, we complement the use of sentence IDs by extracting the raw format that the information appears in the document. For those varied-format fields, we include a complement to the question --- \emph{``and how does it appear in the text?"} --- and change the answers to encompass the canonical-format response (with its respective clue) together with a reserved token \emph{``[text]"} and the raw format of the response as a suffix.
Table~\ref{tab:all_qas} depicts examples of QAs that use the raw format along with the canonical format (column \emph{Raw} marked).

\section{Experimental Setup}
\label{sec:experimental-setup}

We evaluated the effectiveness of the models on four different datasets in Portuguese.
Each dataset represents a specific type of document.
We finetune each model (Section~\ref{sec:models}) on the four datasets (Section~\ref{sec:datasets}), instead of training individually one model for each type of document.
We measure the effectiveness using exact matching (EM), which corresponds to the accuracy; and token-based F1. 

\subsection{Models}
\label{sec:models}

In this section we describe the pretrained and finetuned models used in this work. Some of those models were taken from the shelf, while for others we adapted to our particular task and domain.

{\noindent\bf T5:} Encoder-decoder model that is a Transformer~\cite{vaswani2017attention} with a combination of relevant insights from a systematic study~\cite{raffel2019exploring}. T5 achieved state-of-the-art results in many NLP benchmarks. We work with texts in Portuguese, so this English model acts as a weak baseline.

{\noindent\bf PTT5:} It consists of a T5 model, without changes in the architecture, pretrained on a large Brazilian Portuguese corpus~\cite{carmo2020ptt5}. 
We employ PTT5 Base which showed to be not only more efficient but also more effective than the PTT5 Large~\cite{carmo2020ptt5}. 

{\noindent\bf PTT5-Legal:} To adapt PTT5 to in-domain text, we further pretrain it on a large corpus that comprises two out of the four document types we explore in this work: registries of property and legal publications. This corpus contains approximately 50 million tokens, which is a fraction of 2.7 billion tokens used to pretrain PTT5.

{\noindent\bf PTT5-QA:} We also intend to improve the ability of the model to extract information from diverse documents by question-answering, regardless of the domain of the data. For that, we finetune PTT5 on the SQuAD v1.1 dataset~\cite{rajpurkar2016squad} before finetuning it on our datasets.

{\noindent\bf PTT5-Legal-QA:} Finally, we investigate if the unsupervised in-domain pretraining followed by supervised task-aware finetuning provides effectiveness gains on the task of extracting relevant information from documents.

\subsection{Datasets}
\label{sec:datasets}

We finetune and evaluate the models on four datasets that contain legal documents in Portuguese.
Table~\ref{tab:statistics} presents statistics of the datasets.\footnote[1]{As the legal and registration documents are of private nature, the datasets cannot be made available.}

{\noindent\bf NM-Property:} is composed by the opening section of property documents that discriminate a property or land, containing essential information for legal identification, such as location or qualification of the owners (if an individual or legal entity). 
The documents are scanned and susceptible to OCR errors due to poor image quality.
The dataset has 3191 documents for training, 799 for validation and 242 for testing.

{\noindent\bf NM-Certificates:} represents a set of documents issued by the competent agency of the tax office whose function is to prove that a person, organization or property has or does not have debts, that is, that there are no collection of actions in relation to this natural or legal person, nor in relation to real estate.
The dataset has 760 documents for training, 191 for validation and 311 for testing.

{\noindent\bf NM-Publications:} is comprised of documents issued by the Brazilian government gazette (\textit{Diario Oficial}, similar to U.S. Federal Register). The documents contain legal notices that have been authorised or that are required by law to be published.
The dataset has 1600 documents for training, 401 for validation and 500 for testing.

{\noindent\bf NM-Forms:} consists of forms for opening a bank account filled with personal information. In general, most of the fields are filled. The scanned documents present common OCR errors, such as confusing vertical lines in text boxes as characters.
The dataset has 240 documents for training, 60 for validation and 282 for testing.


\begin{table}[!ht]
\caption{Statistics of the datasets.}\label{tab:statistics}
\centering
\begin{tabular}{l rrrrr}
\hline
\textbf{Dataset} & \textbf{Chars/doc} & \textbf{\ \ \ Fields} & \textbf{\ \ \ Train} & \textbf{\ \ \ Valid} & \textbf{\ \ \ Test}\\
\hline
NM-Property & 3011.53 & 17 & 3191 & 799 & 242 \\
NM-Certificates & 4914.39 & 10 & 760 & 191 & 311 \\
NM-Publications & 1895.76 & 3 & 1600 & 401 & 500\\
NM-Forms & 1917.14 & 25 & 240 & 60 & 282 \\
\hline
Total & - & 55 & 5791 & 1451 & 1334 \\
\hline
\end{tabular}
\end{table}

\subsection{Training and Inference}
\label{sec:training_inference}

For input preprocessing, we prepare sliding windows with an overlap of 50\%. Thus, each window --- that can be interspersed by sentence ID tokens or not --- concatenated with the question occupies a maximum of 512 tokens. Regarding the use of sentence IDs, we use line breaks as a criterion to divide the document text into sentences.
We finetune the models using AdamW optimizer with a constant learning rate of 1e-4 and weight decay of 5e-5. Training takes approximately 22,500 optimization steps with batches of 64 samples. 
The model we select for testing is the one with the highest EM over the validation set.
At test time, the information is extracted using beam search with 5 beams. 
The maximum output length is set to 256 tokens, but none of our test examples reached this limit.

\section{Results}
\label{sec:results}



The first set of experiments involves finetuning the pretrained models presented in Section~\ref{sec:models} for information extraction using individual questions and answers as well as the original document text as context. Rows 1 to 6 in Table~\ref{tab:results} present the results. 

\begin{table}[htbp]
\caption{Main results. ``Pre-ft'' means finetuning on the English SQuAD 1.1 dataset before finetuning on the target datasets.}\label{tab:results}
\centering\resizebox{\textwidth}{!}{
\begin{tabular}{llcccc |lrrrrr}
\hline
& & & & & & & \multicolumn{5}{c}{{\bf Dataset}}  \\
& \textbf{Pre-train} & \textbf{Pre-ft} & \textbf{Comp} & \textbf{Sent} & \textbf{Raw\ \ }                                                  & & {\bf \ \ Prop} & {\bf \ \ Cert} & {\bf \ \ Publ} & {\bf \ \ Form} & {\bf \ \ Avg.} \\ \hline
\multirow{2}{*}{(1)} & \multirow{2}{*}{Eng} & \multirow{2}{*}{-} & \multirow{2}{*}{-} & \multirow{2}{*}{-} & \multirow{2}{*}{-}                   
                                      & EM & 57.5 & 39.4 & 64.3 & 69.1 & 57.6 \\
&&&&&                                 & F1 & 79.0 & 73.5 & 86.7 & 89.4 & 82.2 \\ \hline
\multirow{2}{*}{(2)} & \multirow{2}{*}{Port (Eng vocab)} & \multirow{2}{*}{-} & \multirow{2}{*}{-} & \multirow{2}{*}{-} & \multirow{2}{*}{-}      
                                      & EM & 68.2 & 40.2 & 64.4 & 70.9 & 60.9 \\
&&&&&                                 & F1 & 84.3 & 74.1 & 86.6 & 90.3 & 83.8 \\ \hline
\multirow{2}{*}{(3)} & \multirow{2}{*}{Port} & \multirow{2}{*}{-} & \multirow{2}{*}{-} & \multirow{2}{*}{-} & \multirow{2}{*}{-}                  
                                      & EM & 84.1 & 79.9 & 95.9 & 93.8 & 88.4 \\
&&&&&                                 & F1 & 92.2 & 91.4 & 98.0 & 97.2 & 94.7 \\ \hline
\multirow{2}{*}{(4)} & \multirow{2}{*}{Port} & \multirow{2}{*}{\checkmark} & \multirow{2}{*}{-} & \multirow{2}{*}{-} & \multirow{2}{*}{-}         
                                      & EM & 83.6 & 79.2 & 95.9 & 94.2 & 88.2 \\
&&&&&                                 & F1 & 91.8 & 91.3 & 98.0 & 97.4 & 94.6 \\ \hline
\multirow{2}{*}{(5)} & \multirow{2}{*}{Port+Legal} & \multirow{2}{*}{-} & \multirow{2}{*}{-} & \multirow{2}{*}{-} & \multirow{2}{*}{-}            
                                      & EM & 84.5 & 80.5 & 96.1 & 94.1 & 88.8 \\
&&&&&                                 & F1 & 92.3 & 91.8 & 98.1 & 97.3 & 94.9 \\ \hline
\multirow{2}{*}{(6)} & \multirow{2}{*}{Port+Legal} & \multirow{2}{*}{\checkmark} & \multirow{2}{*}{-} & \multirow{2}{*}{-} & \multirow{2}{*}{-}   
                                      & EM & 84.1 & 82.1 & 96.8 & 94.4 & 89.4 \\
&&&&&                                 & F1 & 92.2 & 92.9 & 98.4 & 97.5 & 95.3 \\ \hline
\multirow{2}{*}{(7)} & \multirow{2}{*}{Port+Legal} & \multirow{2}{*}{\checkmark} & \multirow{2}{*}{\checkmark} & \multirow{2}{*}{-} & \multirow{2}{*}{-} 
                                      & EM & 81.7 & 84.4 & 97.2 & 94.8 & 89.5 \\
&&&&&                                 & F1 & 90.8 & 93.4 & 98.6 & 97.6 & 95.1 \\ \hline
\multirow{2}{*}{(8)} & \multirow{2}{*}{Port+Legal} & \multirow{2}{*}{\checkmark} & \multirow{2}{*}{\checkmark} & \multirow{2}{*}{\checkmark} & \multirow{2}{*}{\checkmark} 
                                      & EM & 81.4 & 85.3 & 96.9 & 94.1 & 89.4 \\
&&&&&                                 & F1 & 90.4 & 93.2 & 98.4 & 97.2 & 94.8 \\ \hline
\end{tabular}
}
\end{table}

We experimented with distinct T5 Base models: the original T5 Base with English tokenizer (row 1), with further pretraining on Portuguese corpus~\cite{carmo2020ptt5} but using the English tokenizer (row 2), and PTT5 Base (row 3). We notice that the adoption of a Portuguese tokenizer by PTT5 provided an error reduction in EM of 70.3\% over the previous experiment that used the same pretraining dataset (rows 2 vs 3).

Afterward, we used the PTT5 model finetuned on the question-answering task, and further finetuned it on the four datasets (PTT5-QA) (row 4). We observe that adapting the model for QA on the SQuAD dataset did not provide improvements over the large-scale pretraining. 

We then applied the PTT5 model pretrained over a corpus of legal documents (PTT5-Legal), and finetuned it on the four datasets. The unsupervised pretraining brought the best result over the NM-Properties dataset, and a minor improvement on the average of the four datasets (row 5).
Finally, we explored the PTT5 model that incorporated the finetuning on question-answering after in-domain pretraining (PTT5-Legal-QA). This model achieved the best average EM and F1 (row 6).

\subsection{Experiments for Compound QAs}
\label{sec:res_comp_qas}

In this section, we investigate whether the use of compound QAs, described in Section~\ref{sec:comp_qas}, brings benefits in effectiveness or at least reaches comparable performance to the use of individual QAs. We applied compound QAs for address, land area, private area and built area from NM-Property; and for address from NM-Forms.
To compare results, we post-process the compound answers by splitting them into individual sub-responses, each one including the clue in square brackets and the extracted information. 

As shown in row 7 of Table~\ref{tab:results}, compound QAs yields better results than using individual QAs, reaching superior performance on three datasets. We emphasize, however, that the average results are comparable. This confirms that the use of compound QAs brings efficiency and does not affect the model effectiveness.

However, the EM results for NM-Property decreased from 84.1\% to 81.7\%. Most errors occur for the address field, in scenarios in which the response is N/A (often not annotated), but the model outputs an answer. One of the reasons is that the NM-Property concentrates a large number of registries in rural or allotted areas, in which some information is missing. 
We noticed that the four most negatively impacted fields are from NM-Property's address: complement, street, district and number. On the other hand, five out of seven address subfields of NM-Form had improvements.


\subsection{Experiments for Sentence IDs and Canonical Format}
\label{sec:res_sent_ids_canonical}

Herein, we investigate whether exploring sentence IDs and using the raw format along with the canonical format text described in Section~\ref{sec:sent_ids_canonical}, provides comparable results to the previous set of experiments. 

We employed canonical format extraction for areas, dates, ID numbers, states and certificate results. Specifically, we use it for 8 NM-Property fields, 4 NM-Certificate fields and 7 NM-Forms fields.
To compare results, we post-process the compound answers by splitting them into individual sub-responses. Each sub-response involves the clue in square brackets and the information, preceded by sentence ID in brackets, and succeeded by the text in raw format.
Results in row 8 of Table~\ref{tab:results} show that this method does not outperform the one without sentence IDs and raw-text extractions. 
Nonetheless, the model robustness for extracting information on varied legal and registration documents was not affected.

\subsection{Comparison with BERT on a NER task}
\label{sec:ablation}

In this experiment, we compare the extraction capabilities of our proposed framework to a named entity recognition (NER) system, a common part of classical IE pipelines. We use the NER implementation of Souza et al.~\cite{souza2020bertimbau}, which uses BERT and casts NER as a token-level classification task.
Since NER models cannot predict nested or overlapping entities and requires all outputs to be tied to input tokens, we filter out all classification fields and commonly overlapping field classes from our dataset. We also remove entire documents if they contain overlapping entities of the remaining fields. The reduced dataset retains 38\% of the documents and 42 of the 55 fields. We then finetune both the NER and the PTT5-Legal-QA models on the reduced training set and evaluate on the reduced test set. 

Following CoNLL03 evaluation procedures, the evaluation is performed at entity-level using exact match. For a given entity, the prediction is marked as correct only if the system predicts the correct entity label and location (start and end offsets) in the input text. Thus, this experiment validates exclusively the performance of our output alignment method detailed in \ref{sec:res_sent_ids_canonical}.

\vspace{-0.5cm}
\begin{table}[!ht]
\caption{NER ablation results.}\label{tab:ner_ablation_results}
\vspace{-0.1cm}

\centering
\begin{tabular}{l | c | c | c | c}
\hline
  Model & Params & Precision & Recall & F1-score (micro) \\ \hline
  BERT-Large & 330M & 90.2 & 92.9 & 91.5 \\
  T5-base (ours) & 220M & 91.6 & 89.6 & 90.6 \\ \hline
  
\end{tabular}
\end{table}

Table \ref{tab:ner_ablation_results} shows the results. Our seq2seq framework shows a slightly lower but competitive performance on the task, even though this ablation setup gives an edge to the NER system by removing impossible cases for NER, in which it would be penalized with false negatives. This task also does not evaluate answer post-processing, which is one of seq2seq's strengths and would have to be done by an extra component. 

\section{Conclusion}
\label{sec:conclusion}
We validated the use of a single seq2seq model for extracting information from four different types of legal and registration documents in Portuguese.
The model is trained end-to-end to output structured text, thus replacing parts of rule-based normalization and post-processing steps of a classical pipeline. The compound QAs also replaces a relationship extraction step that is required when fields are extracted independently in classical pipelines, such as by a named entity recognition model.
We found that target language (Portuguese) pretraining and tokenization are the most important adaptations to increase effectiveness, while pretraining on in-domain (legal) texts and finetuning on a large question-answering dataset marginally improve results.
Finally, we propose a method to align answers with the input text, thus allowing seq2seq models to be more easily monitored and audited in IE pipelines.




\bibliographystyle{splncs04}
\bibliography{mybibliography}

\end{document}